\pdfoutput=1
\documentclass[11pt]{article}

\usepackage[final]{acl}

\usepackage{times}
\usepackage{latexsym}
\usepackage{enumitem}
\setlist{itemsep=1pt, topsep=1pt}
\usepackage{booktabs}
\usepackage{makecell}
\usepackage{rotating}

\usepackage[T1]{fontenc}
\usepackage[utf8]{inputenc}

\usepackage{microtype}

\usepackage{inconsolata}

\usepackage{graphicx}

\title{Social Bias in Multilingual Language Models: A Survey}

\author{Lance Calvin Lim Gamboa\textsuperscript{1,2} \quad
 Yue Feng\textsuperscript{1} \thanks{Corresponding author} \quad Mark Lee\textsuperscript{1}\\
 \textsuperscript{1}School of Computer Science, University of Birmingham\\
 \textsuperscript{2}Department of Information Systems and Computer Science, Ateneo de Manila University\\
 \\
  {\texttt{llg302@student.bham.ac.uk}, \texttt{lancecalvingamboa@gmail.com}} \\
  {\texttt{y.feng.6@bham.ac.uk}} \\
  {\texttt{m.g.lee@bham.ac.uk}} \\}

\begin{document}
\maketitle
\begin{abstract}
Pretrained multilingual models exhibit the same social bias as models processing English texts. This systematic review analyzes emerging research that extends bias evaluation and mitigation approaches into multilingual and non-English contexts. We examine these studies with respect to linguistic diversity, cultural awareness, and their choice of evaluation metrics and mitigation techniques. Our survey illuminates gaps in the field’s dominant methodological design choices (e.g., preference for certain languages, scarcity of multilingual mitigation experiments) while cataloging common issues encountered and solutions implemented in adapting bias benchmarks across languages and cultures. Drawing from the implications of our findings, we chart directions for future research that can reinforce the multilingual bias literature’s inclusivity, cross-cultural appropriateness, and alignment with state-of-the-art NLP advancements.
\end{abstract}

\section{Introduction}
Multilingualism has grown to be a core property of recently released pretrained language models (PLMs), such as GPT-4 \citep{openai2023gpt}, Llama 3 \citep{meta2024llama3}, and Qwen 2 \citep{yang2024qwen2technicalreport}. The model release reports of these models all include evaluations on multilingual language understanding benchmarks and demonstrate the models’ remarkable performances on these tests. These models’ multilingual capabilities have been confirmed by independent assessments done by NLP researchers, such as \citet{zhao2024llamaenglishempiricalstudy} and \citet{huang-etal-2023-languages}. Concurrently, there are also emerging endeavors to create models that specialize on handling tasks in multiple languages—e.g., Aya \citep{ustun-etal-2024-aya} and BLOOMZ \citep{muennighoff-etal-2023-crosslingual}—or a specific non-English language—e.g., HyperCLOVA X for Korean \citep{yoo2024hyperclovaxtechnicalreport}, ChatGLM for Chinese \citep{glm2024chatglmfamilylargelanguage}, and Vietcuna for Vietnamese \citep{vilm2023vietcuna}. 

\begin{figure}[!b]
\centering
  \includegraphics[width=0.75\columnwidth]{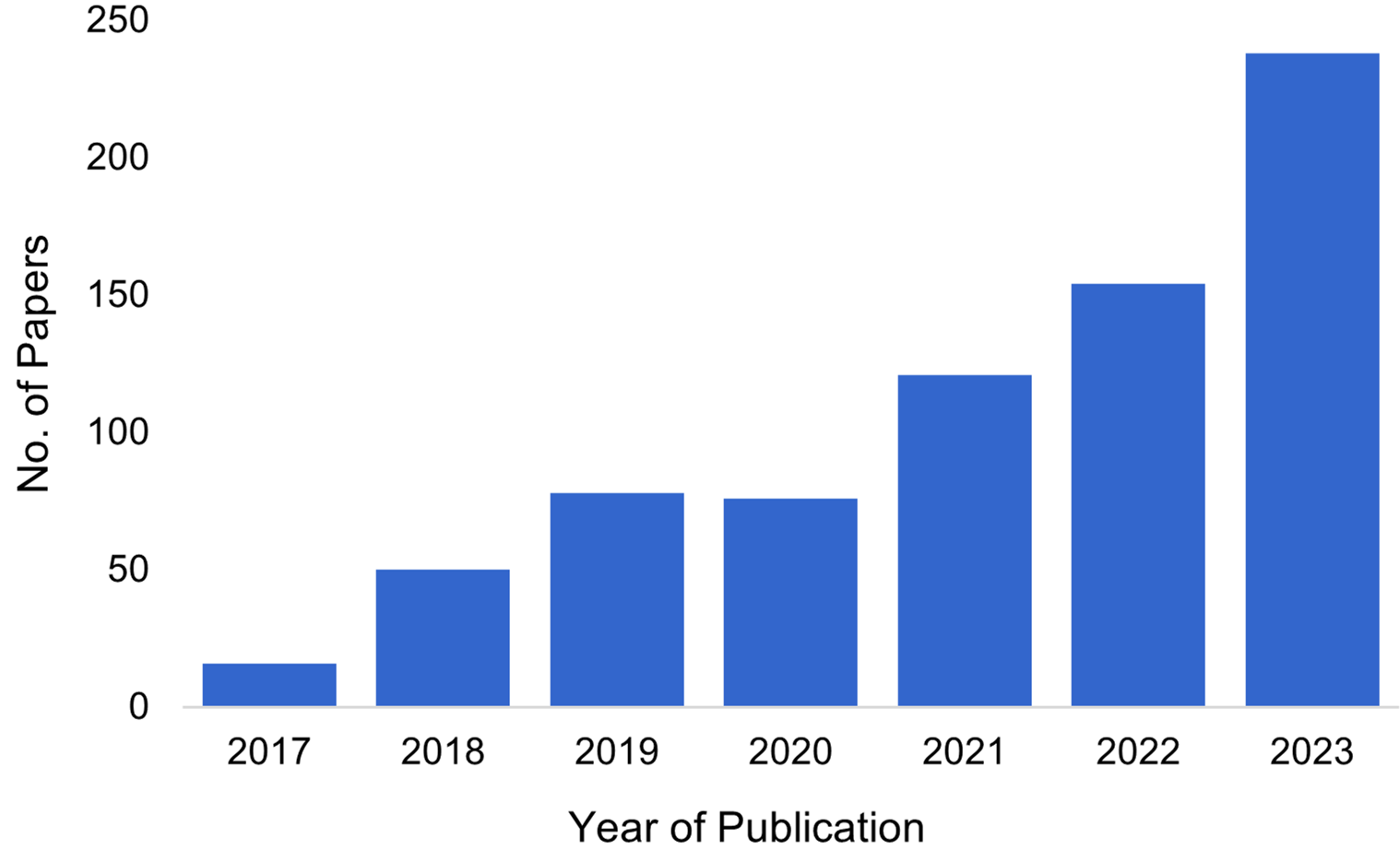}
  \caption{Number of ACL Anthology papers that contain the terms \textit{bias}, \textit{harm}, \textit{stereotype}, \textit{toxic}, or \textit{fair} and \textit{multilingual}, \textit{cross-lingual}, \textit{interlingual}, or a non-English language in the title or abstract.}
  \label{fig:year_of_pub}
\end{figure}

Multilingual models are not exempt from the safety and bias issues that have been identified in models handling English. Pioneering studies calling attention to the biased behaviors of English models (e.g., \citealp{caliskan2016weat}; \citealp{nangia2020crows}) have been followed by replications demonstrating the presence of similar problems in models processing non-English languages (e.g., \citealp{lauscher-etal-2020-araweat}; \citealp{neveol-etal-2022-french}). As such, NLP scholars around the world have progressively expanded efforts to evaluate and ensure the fairness of multilingual and non-English models. This increasing efforts are demonstrated by Figure \ref{fig:year_of_pub}, which depicts the rising number of papers in this niche. 

While these multilingual bias studies employ an eclectic selection of approaches, many utilize methods that have been criticized as being error-prone and culturally unaware—e.g., simply relying on automated translations in adapting English bias tests to non-English languages \citep{talat2022reap}. These practices are concerning as they may lead not only to the underestimation of culturally specific biases within PLMs but also to a focus on Anglocentric concepts of fairness in the field of bias evaluation and mitigation. There thus stands a need to take stock of the multilingual PLM bias literature, consider the approaches it has been using, and check how successfully such approaches have been making multilingual models more inclusive. While multiple surveys of the general PLM bias scholarship have already been conducted (e.g., \citealp{gallegos-etal-2024-bias}; \citealp{gupta-etal-2024-sociodemographic}; \citealp{goldfarb-tarrant-etal-2023-prompt}; \citealp{10.1145/3597307}), most just list the investigation of non-English biases as a direction for future research and consequently fail to engage the growing number of studies in this area. 

In this paper, we address this gap by systematically and critically reviewing studies on multilingual and non-English PLM bias. To identify these papers, we applied a systematic keyword-based search on ACL Anthology, IEEE Xplore, and the proceedings of the NeurIPS, FAccT, and AIES conferences. From these databases, we shortlisted NLP and language modeling articles that included the following strings in their titles or abstracts: \textit{bias}, \textit{fair}, \textit{toxic}, \textit{harm}, or \textit{stereotyp} plus \textit{multilingual}, \textit{cross-lingual}, \textit{interlingual}, \textit{multiple languages}, or any language name enumerated in ISO 639 or the Codes for the Representation of Names of Languages \citep{libraryofcongress2017iso}. We then constrained our selection to papers published on or before December 31, 2024. We also filtered out articles that fulfilled the above criteria but did not sustantially engage the concept of sociodemographic bias (e.g., papers about statistical, inductive, and positional bias). This process resulted in a final article set consisting of 106 articles relevant to multilingual PLM bias—97 from ACL Anthology, 7 from IEEE Xplore, 1 from FAccT, and 1 from AIES. 

We examine these works using an annotation taxonomy we developed. Our taxonomy builds on extant PLM bias typologies (i.e., \citealp{gallegos-etal-2024-bias}; \citealp{gupta-etal-2024-sociodemographic}; \citealp{goldfarb-tarrant-etal-2023-prompt}) and extends these with categories relating to choice of languages and language families, dataset adaptation methods, and cultural awareness in methodological design. By clarifying these aspects among studies we inspected, our taxonomy allows us to explore issues in devising bias evaluation and mitigation protocols for multicultural contexts. Concurrently, we also highlight solutions that have been taken to navigate these concerns. 

Our survey exposes the multilingual PLM bias literature’s preference for Chinese, Indo-European, and other highly resourced languages. This partiality leads to a shortage of bias research on languages spoken by major sections of the global population and by cultures most active in the adoption of AI technologies. We also discover that more than half of non-English bias tests are accompanied by methodological protocols which do not explicitly document cultural considerations in the benchmark development process. This lack of transparency makes it unclear how (or if) these benchmarks engaged with the adaptation issues confronted by their more culturally aware counterparts—for example, the generalizability of some bias dimensions (e.g., race), the localization of universal biases, and the resolution of differences in linguistic gender across languages. Finally, our review also reveals that multilingual bias research largely stops at the evaluation stage and rarely crosses into the mitigation of biases. This finding highlights the urgency of developing debiasing approaches for multilingual models or, at least, of inspecting the applicability of English debiasing methods on non-English contexts.

Our contributions are threefold: 
\setlength{\itemsep}{0pt} 
\setlength{\topsep}{0pt} 
\begin{itemize}
    \item We synthesize works on multilingual bias and pinpoint gaps and best practices in the field, thereby revealing and encouraging reflection about trends in the literature.
    \item Our review sheds light on common challenges encountered by multilingual bias researchers and the steps they have taken to solve these. This catalog of challenges and solutions can guide the design of future work in the field.
    \item We compile a concise agenda for future multilingual bias research based on issues and limitations we identified in our review.
\end{itemize}

The rest of this paper is structured as follows: we briefly describe our method for systematic review, particularly how we selected papers and examined them using our taxonomy (\ref{sec:method}). Next, we outline our findings and their implications, starting with our observations on the linguistic diversity of the PLM bias literature (\ref{sec:linguistic_diversity}) and the cultural awareness of methods used to broaden this diversity (\ref{sec:cultural_awareness}). We continue with a review of evaluation and mitigation techniques applied on multilingual and non-English models (\ref{sec:eval_mitig}). We conclude with a list of opportunities to improve future research in the field (\ref{sec:conclusion}). 

\section{Annotation}
\label{sec:method}

%\subsection{Bias Statement}
We follow the conceptualization of social bias utilized by the survey papers of \citet{gupta-etal-2024-sociodemographic} and \citet{gallegos-etal-2024-bias}, who define PLM bias as inequalities in how PLMs generate outcomes or perform when handling data and inputs associated with diverse social demographics. We also recognize the distinction between biases leading to representational harms and those leading to allocational harms \citep{crawford2017trouble,barocas2017problem}. The former occur when PLMs propagate stereotypes, toxic language, and disparate judgments that depict one social group more unfavorably than another \citep{blodgett-etal-2020-language, crawford2017trouble}, while the latter emerge from PLMs distributing resources or opportunities unfairly across groups \citep{blodgett-etal-2020-language,barocas2017problem}. However, because most non-English languages lack labeled pretraining data and therefore have few to no predictive NLP systems that allocate resources \citep{joshi-etal-2020-state}, the studies we scrutinize only analyze representationally harmful biases. 

Given this conceptualization of bias we adopted, we took the bias dataset annotation scheme developed by \citet{goldfarb-tarrant-etal-2023-prompt} as a starting point in developing our own taxonomy. Rooted in an understanding of the potential representation harms of PLMs, their taxonomy notes (1) basic scope attributes about a paper and its accompanying dataset/s (e.g., language/s used, model/s tested, code availability) and (2) aspects about how a paper operationalizes bias evaluation (e.g., bias metric/s, benchmark format, proxies for demographic groups). We then refined the taxonomy with the typologies proposed by \citet{gallegos-etal-2024-bias} and \citet{gupta-etal-2024-sociodemographic}, whose definitions of bias we also utilize. This led to a revision of the categories used to classify bias metrics and benchmark entries and the addition of a mitigation-related annotation attribute. We also leveraged our familiarity with the field of multilingual PLM bias to augment the taxonomy with elements relating to the originality of the non-English benchmarks, the benchmark development method, and the cultural nuances involved therein. Applying this initial taxonomy on the articles and revising it based on new categories and labels that emanated from the literature resulted in the final taxonomy in Appendix \ref{app:taxonomy}. 

\begin{figure*}[!ht]
\centering
  \includegraphics[width=0.8\textwidth]{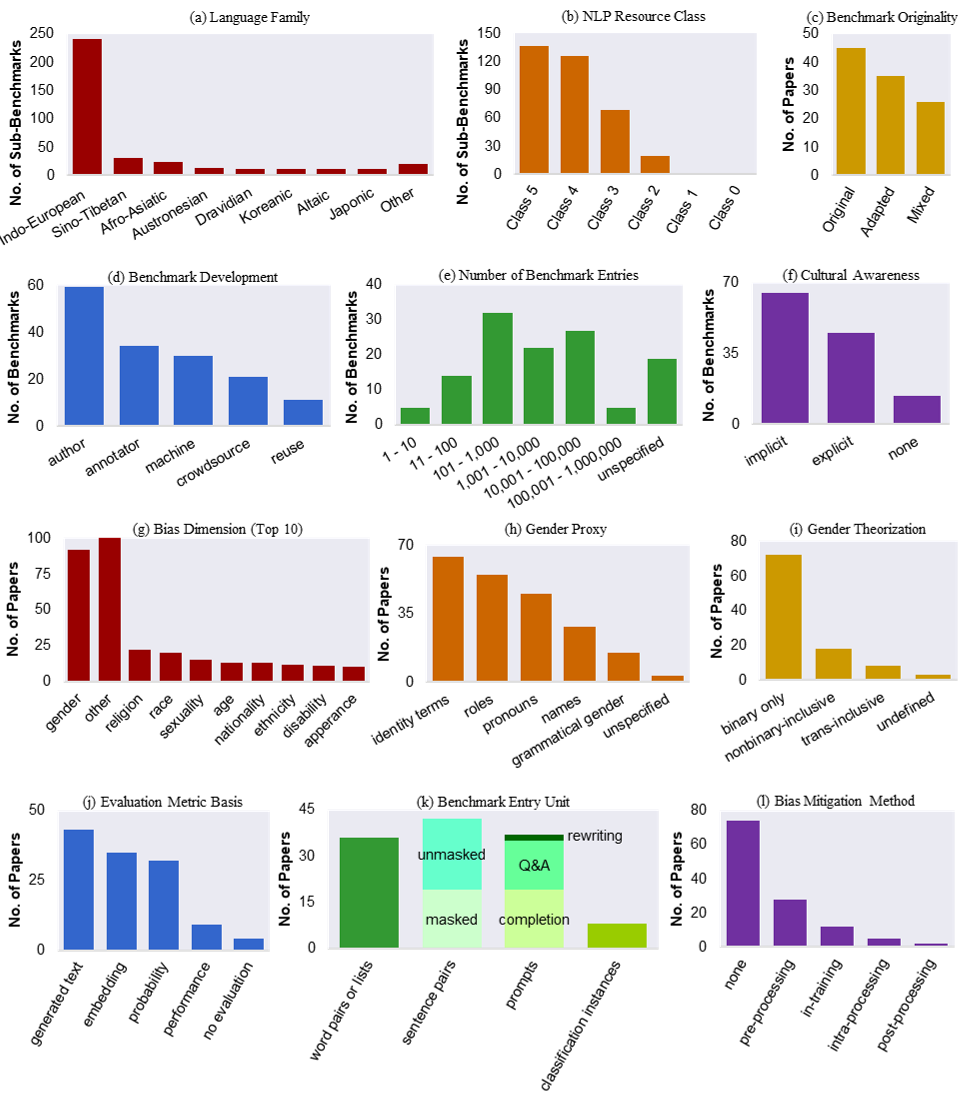}
  \caption{Results for annotating 106 multilingual bias articles using our taxonomy.}
  \label{fig:findings}
\end{figure*}

The authors of this paper used this taxonomy to conduct annotations. Disagreements were infrequent and labeling was straightforward. We release\footnote{\url{https://github.com/gamboalance/multilingual_bias_survey}} a consolidated list of the papers we examined and our annotations for each paper. Figure \ref{fig:findings} illustrates a quantitative summary of our annotation and analysis, which we expound upon in the next three sections. 

\section{Language Choice and Diversity}
\label{sec:linguistic_diversity}
Several of the studies we examined used more than one benchmark; therefore, we looked into a total of 124 bias benchmarks in this survey. Among these benchmarks, we further disaggregated multilingual ones into monolingual sub-benchmarks, resulting in a total of 376 single-language sub-benchmarks analyzed for this paper. All in all, the bias tests we annotated featured 67 different languages (listed in Appendix \ref{app:languages}), with Chinese taking the top spot in terms of frequency ($n=30; 7.98\%$), followed by Spanish ($n=28, 7.45\%$), French ($n=24, 6.38\%$), German ($n=24, 6.38\%$), and Arabic ($n=20, 5.32\%$). As illustrated in Figures \ref{fig:findings}(a) and \ref{fig:findings}(b), grouping the languages into their language families and into NLP resource classes reveals that \textbf{the literature has an asymmetric focus on Indo-European languages} ($n=242, 64.36\%$) \textbf{and on languages that \citet{joshi-etal-2020-state} would classify as highly resourced in NLP} ($n_{Class 5}=137, 36.44\%$; $n_{Class 4}=126, 33.51\%$). 

These disproportionate imbalances show that there is a \textit{linguistic bias} in multilingual PLM bias research. Next to English, most PLM bias studies address problematic model behavior mainly in languages spoken by economically developed countries (as determined by GDP per capita data from \citealp{imf2024gdp}). While bias studies in these languages are undoubtedly important, they are unable to address the negative repercussions of AI being used enthusiastically in less developed countries, like India, Indonesia, and the Philippines \citep{sarkar2023aiindustry}. Consequently, the statistics above lend empirical support to observations that the communities governing and regulating language model development and moderation are removed from the majority of the communities using these technologies \citep{talat2022reap}. 

Furthermore, 5 of the 35 most widely spoken languages in the world \citep{ethnologue2023} have $<10$ bias tests written in their languages (e.g., Bengali, Indonesian), while 9 have $<5$ tests (e.g., Swahili, Persian) and 5 more are completely absent in the bias literature (e.g., Hausa, Javanese). This pattern in the literature risks values in only a limited number of cultures (i.e., white, Western, or Chinese) being represented in endeavors to train and safeguard NLP systems  (as previously found by \citealp{kreutzer2022} and \citealp{thysltrup2020}). Despite conveying an impression of linguistic inclusivity in the PLM bias literature becoming, this \textit{linguistic bias}—if left unchecked—may end up obscuring culturally specific issues in models and exacerbating inequities in what communities and perspectives are considered in the field. We therefore agree with appeals to empower technologically marginalized agents in contributing towards efforts to develop responsible AI \citep{talat2022reap}. 

\section{Cultural Awareness in Benchmark Development}
\label{sec:cultural_awareness}
Figure \ref{fig:findings}(c) shows that the number of works that develop their own benchmark ($n=45, 42.45\%$) is almost comparable to the number of papers that adapted a pre-existing bias test to their chosen language/s ($n=35, 33.02\%$). Examples of adapted benchmarks are Multilingual HolisticBias \citep{costa-jussa-etal-2023-multilingual} and AraWEAT \citep{lauscher-etal-2020-araweat}, which were created through the translation of American-sourced stereotypes found in English HolisticBias \citep{smith2022imsorry} and English WEAT \citep{caliskan2016weat} respectively. In contrast, original benchmarks include CHBias \citep{zhao-etal-2023-chbias} and KoSBi \citep{lee-etal-2023-kosbi}, the authors of which collected novel prejudices relevant to Chinese and Korean societies. A minority of studies employ a mix of original and adapted benchmarks ($n=26, 24.53\%$)—e.g., the French CrowS-Pairs testbed \citep{neveol-etal-2022-french}, which is composed of entries translated from the original CrowS-Pairs \citep{nangia2020crows} and new sentences sourced from French contributors. %Some, like the creators of CBBQ \citep{huang-xiong-2024-cbbq} and KoBBQ \citep{jin-etal-2024-kobbq}, adapted an extant benchmark development framework (i.e., the BBQ framework by \citealp{parrish2022bbq}) in order to collect new biases and prompts.

Authors often take the lead in creating or adapting entries for non-English tests, with about half of the benchmarks ($n=59, 47.58\%$) having some significant authorial contribution in their development—as demonstrated in Figure \ref{fig:findings}(d). Such contribution often comes in the form of the authors manually translating English entries \citep{fort-etal-2024-stereotypical,gamboa2023evaluating} or personally constructing culturally appropriate prompts \citep{wang-etal-2024-chinese,ibaraki-etal-2024-analyzing}. In writing the latter, they relied on stereotypes mined from pre-existing corpora (e.g., Wikidata and CommonCrawl in \citealp{naous-etal-2024-beer}), mass and social media \citep{zhu-etal-2024-plms,huang-xiong-2024-cbbq}, academic articles \citep{gamboa2023characterizing}, or government documents and statistics (e.g., job market data in \citealp{steinnun-fridriksdottir-einarsson-2024-gendered}; state-owned name databases in \citealp{das-etal-2023-toward}). 

Author-driven benchmark development, however, has been criticized for lacking diversity of perspective because of authors’ limited familiarity with some biases \citep{goldfarb-tarrant-etal-2023-prompt}. To remedy this issue, a number of studies have employed cultural insiders—namely annotators or experts ($n=34, 27.42\%$) and crowdsource workers ($n=21, 16.94\%$)—to gather a multiplicity of viewpoints in creating their datasets. These cultural insiders helped in either writing the benchmark entries (e.g., \citealp{huang-xiong-2024-cbbq}; \citealp{touileb-nozza-2022-measuring}) or validating translations and stereotypes provided by the authors, other annotators, and pre-existing benchmarks (e.g., \citealp{grigoreva-etal-2024-rubia}; \citealp{mukherjee-etal-2023-global}). For quite a few of the studies though, relying solely on humans was insufficient as they aimed to create benchmarks with entries numbering in the ten- and hundred-thousands—as shown in Figure \ref{fig:findings}(e). As such, they used NLP models to aid in their evaluation of PLMs ($n = 30, 24.19\%$). Some used translation technologies to adapt benchmarks into another language (e.g., NLLB and Google Translate in \citealp{sahoo-etal-2024-indibias}), others leveraged generative PLMs to write prompts or stereotypes (e.g., \citealp{huang-xiong-2024-cbbq}), and several used algorithms to automatically populate templates and create a large number of test items (e.g., \citealp{jin-etal-2024-kobbq}). 

Unfortunately, \textbf{no matter the benchmark development method, most works do not explicitly document the cultural nuances considered in creating bias tests}. As seen in Figure \ref{fig:findings}(f), about a tenth of the benchmarks examined ($n=14, 11.29\%$) were created with no attention paid to pertinent cultural considerations; meanwhile, more than half ($n=65, 52.42\%$) only implied a semblance of cultural awareness through the involvement of cultural insiders (e.g., annotators, experts, crowdsource workers) but did not record what adaptation issues and processes these participants engaged with. 

This inclination to overlook aspects of benchmarks development linked to cultural interpretation and calibration is alarming, especially given the number of non-English datasets containing adapted elements ($n=61, 57.55\%$). Nebulous or non-existent descriptions of cultural concerns encountered in multilingual bias test construction make it impossible to assess how appropriate test items are in capturing biases in a particular culture. After all, biases and stereotypes are culture-dependent, and what might be a discriminatory statement in one societal context may be less significant in another \citep{gallegos-etal-2024-bias,talat2022reap}. The lack of clarity among many multilingual bias studies on how they handle these cultural idiosyncrasies casts doubt on the validity of results and conclusions drawn from their culturally naïve benchmarks. There is thus a need to rectify this inadequacy in transparent cultural awareness in the PLM bias literature. 

%For example, one CrowS-Pairs entry measures gender bias by comparing the likelihood that a model would generate the sentences \textit{Cathy/Mark went shoe shopping at the mall} \citep{nangia2020crows}. It may be presumed that this entry might be based on the stereotype that fashion is more strongly associated with one gender; however, such association may only be relevant to the American culture from which CrowS-Pairs arises and is less applicable to European societies, like France \citep{neveol-etal-2022-french}. 

Fortunately, there is a non-negligible amount of research in the field ($n = 45, 36.29\%$) that is forthright in its cultural awareness and that can serve as basis for improving cultural transparency. We reviewed these works and identified common adaptation issues encountered by multilingual bias scholars and the solutions they implemented. We discuss these in the succeeding section and provide a quick summary at Table \ref{tab:adaptation}.

\begin{table*}[!htb]
    \centering
    \footnotesize
    \begin{tabular}{p{3.6cm} p{5.5cm} p{5.5cm}}
        \hline\hline
         \textbf{Paper/s} & \textbf{Adaptation Issue} & \textbf{Adaptation Practice}\\
         \hline\hline
         \multicolumn{3}{c}{\textbf{General Adaptation Practices}}\\ \hline
         \citet{fort-etal-2024-stereotypical,gamboa2023evaluating} & & Authors manually translate entries from English benchmarks into other languages.\\ \hline
         \citet{wang-etal-2024-chinese,ibaraki-etal-2024-analyzing} &  & Authors construct entirely new prompts appropriate to their target culture, relying on the following to determine contextually relevant stereotypes:\\
         \citet{naous-etal-2024-beer} &  & • Wikidata and CommonCrawl\\
         \citet{zhu-etal-2024-plms,huang-xiong-2024-cbbq} &  & • mass and social media\\
         \citet{gamboa2023characterizing} &  & • academic articles\\
         \citet{steinnun-fridriksdottir-einarsson-2024-gendered, das-etal-2023-toward} &  & • government documents and statistics\\ \hline
         \citet{huang-xiong-2024-cbbq,touileb-nozza-2022-measuring} &  & Recruit cultural insiders (e.g., annotators, experts, or crowdsource workers) to write benchmark entries.\\ \hline
         \citet{grigoreva-etal-2024-rubia,mukherjee-etal-2023-global} &  & Recruit cultural insiders to validate translations and stereotypes from the authors, other annotators, or existing benchmarks.\\ \hline
         \citet{sahoo-etal-2024-indibias} &  & Use machine translators to adapt benchmarks into another language.\\ \hline
         \citet{huang-xiong-2024-cbbq} &  & Use generative models to write new prompts or find culturally appropriate stereotypes.\\
         \hline\hline
         \multicolumn{3}{c}{\textbf{Adaptation Practices Pertaining to Bias Dimensions}}\\ \hline
         \citet{devinney-etal-2024-dont,jin-etal-2024-kobbq} & Some bias dimensions in English benchmarks (e.g., racism) are irrelevant to non-American or non-Western contexts. & Streamline race-, ethnicity-, and nationality-related biases into one bias dimension. \\ \hline
         \citet{sahoo-etal-2024-indibias,bhatt-etal-2022-contextualizing,malik-etal-2022-socially,huang-xiong-2024-cbbq,lee-etal-2023-kosbi} & Bias dimensions relevant to specific cultures are absent in English benchmarks. & Add culturally specific bias dimensions (e.g., those related to caste, disease, and family structure) into the adapted benchmark. \\ 
         \hline\hline
         \multicolumn{3}{c}{\textbf{Adaptation Practices Pertaining to Contextualizing Universal Biases}}\\ \hline
         \citet{jin-etal-2024-kobbq, neveol-etal-2022-french} & Some terms in English benchmarks are specfic to Western or American culture (e.g., \textit{rugby} and \textit{star quarterback} referencing sports prominent to the USA). & Replace these terms with contextually compatible equivalents (e.g., using \textit{basketball} instead of \textit{rugby} in cultures where the latter is not popular).  If there are no equivalents, remove the entry containing the culturally irrelevant term.\\ \hline
         \citet{marinova-etal-2023-transformer,10.1145/3630106.3659017} &  & Design completely novel evaluation frameworks and benchmarks based on how biases manifest locally. \\
         \hline\hline
         \multicolumn{3}{c}{\textbf{Adaptation Practices Pertaining to Gender and Sexuality}}\\ \hline
         \citet{steinborn2022information, sahoo-etal-2024-indibias} & Benchmarks relying on counterfactual inputs are difficult to adapt into languages with gender-neutral pronouns and vocabularies (e.g., Finnish). & Translate pronouns into gendered names (e.g., John, Mary) or identity terms (e.g., man, woman). \\ \hline
         \citet{neveol-etal-2022-french} & In heavily gendered languages (e.g., French), gender inflections transform counterfactual pairs into practically different sentences. & Paraphrase the prompts to reduce the need for gender inflections while preserving meaning. \\ \hline
         \citet{chavez-mulsa-spanakis-2020-evaluating,grigoreva-etal-2024-rubia,matthews-etal-2021-gender} & Some gendered words carry multiple meanings, making it impossible to disentangle whether bias effects arise from their inherent gender or from their other meanings. & Eliminate or find substitutes for these words. Alternatively, retain these words but warn readers and benchmark users of their potential impact on evaluation results.\\ 
         \hline\hline
    \end{tabular}
    \caption{Benchmark adaptation practices utilized by multilingual bias researchers.}
    \label{tab:adaptation}
\end{table*}

\section{Issues in Adapting Bias Benchmarks}
\subsection{Bias Dimensions and Their Cultural Relevance}
Across the studies we annotated, we identified 24 social dimensions (Table \ref{tab:bias_dimensions} in Appendix \ref{app:sample_papers}) along which multilingual benchmarks measured bias. Among these attributes, a few consistently appeared in English benchmarks but were deemed by non-American scholars to be irrelevant to their contexts. Racism, in particular, was deemed to be an issue more central to primarily English-speaking countries and was therefore merged with ethnicity- and nationality-based bias in studies conducted in Sweden \citep{devinney-etal-2024-dont} and South Korea \citep{jin-etal-2024-kobbq}. 

\textbf{Other benchmark developers note that dimensions often included in English benchmarks do not capture the totality of biases present in their home societies}. Caste-related stereotypes, for example, are highly salient in Indian society but are never featured in English bias datasets \citep{sahoo-etal-2024-indibias,bhatt-etal-2022-contextualizing,malik-etal-2022-socially}. Other culturally specific dimensions include disease, household registration (relevant to Chinese culture according to \citealp{huang-xiong-2024-cbbq}), pregnancy, family structure, and marital status (relevant to Korean Culture according to \citealp{lee-etal-2023-kosbi}). In future, authors working on multilingual bias evaluation should therefore reflect on how suitable their benchmarks’ dimensions are to the culture of the language they are working with. Furthermore, more efforts should be invested on non-gender-related biases. Sexism is the subject of the vast majority of multilingual bias studies ($n = 92, 86.79\%$), as shown in Figure \ref{fig:findings}(g), and leaves many other types of biases, including intersectional ones, underexplored.

\subsection{Contextualizing Universal Biases}
While some overarching categories of bias cut across cross-cultural boundaries (e.g., gender), their manifestations vary in different localities. A difficulty constantly raised by the examined works is the appearance of culturally specific terms and stereotypes in English benchmarks. For example, although stereotypes between physical activity and gender are prominent worldwide, the way these are expressed in English tests through terms related to American sports culture (e.g., \textit{rugby}, \textit{star quarterback}) may not be apt in some cultures \citep{sahoo-etal-2024-indibias,jin-etal-2024-kobbq}. In adapting entries containing such terms, authors and annotators used their knowledge of their culture to pick a contextually compatible equivalent—e.g., replacing \textit{rugby} with \textit{basketball} in a Korean dataset \citep{jin-etal-2024-kobbq}. 

In some cases, this practice of localization was impossible because a concept or a stereotype itself did not exist in the target culture, compelling developers to just discard these inputs from the adapted benchmark. To demonstrate: stereotypes linking queerness to the color pink and to culinary ability were deemed untranslatable to the French culture and removed from French CrowS-Pairs \citep{neveol-etal-2022-french}. 

Some have gone beyond mere entry rewriting or removal and have intentionally designed their evaluation frameworks with local manifestations of universal biases in mind. For example, \citet{marinova-etal-2023-transformer} drew from their knowledge of the peculiar ethnic composition of Bulgaria’s population to write original prompts assessing masked models’ biases regarding these minorities. Meanwhile, 
\citet{10.1145/3630106.3659017} organized community workshops with women in rural India to collect sentences illustrating Hindi concepts of gender bias. While the first approach yielded findings about how differently language models treated particular Bulgarian ethnic groups, the latter unveiled how Hindi-speaking communities associated the male with curiosity and the female with reservedness. 

The insights and issues surfaced by these contextualization methods affirm the value of culturally aware and transparent benchmark adaptation techniques. \textbf{Without sufficient documentation on the cultural specificities of the adaptation process, it cannot be known if local stereotypes were incorporated in the bias test or if, at the very least, contextually trivial biases were addressed}. 

\subsection{Linguistic Gender and Non-binary Representation}
Figure \ref{fig:findings}(h) illustrates that multilingual bias studies use different proxies to denote gender in their benchmark entries. The use of some of these proxies, however, comes with challenges in multilingual research because of differences in how linguistic gender is expressed across different languages.

One such challenge is the gender neutrality of some languages leading to the homogenization of counterfactual inputs that many benchmarks rely on. \citet{steinborn2022information}, for example, needed to deal with the Finnish language having only the genderless third person pronoun \textit{hän}. Such non-gendered-ness transformed originally different benchmark entry pairs (e.g., \textit{\textbf{He} was timid.} / \textit{\textbf{She} was timid.}) into two identical sentences (e.g., \textit{\textbf{Hän} oli arka.} / \textit{\textbf{Hän} oli arka.} in Finnish), rendering it impossible for bias metrics to compare how differently a PLM would behave with respect to each gender. As a solution, pronouns in the original benchmarks were changed to either identity terms \citep{sahoo-etal-2024-indibias} or gendered names \citep{steinborn2022information}. In some cases, entries involving gender-neutral pronouns were removed from the adapted benchmark (e.g., \citealp{ousidhoum-etal-2021-probing}; \citealp{matthews-etal-2021-gender}).

\begin{table*}[!htb]
    \centering
    \small
    \begin{tabular}{p{4.7cm} p{3.9cm} p{6cm}}
        \hline\hline
         \textbf{Evauation Metric Category} & \textbf{Benchmark Entry Units} & \textbf{Sample Papers}\\
         \hline\hline
         \multicolumn{2}{l}{\textbf{Embedding-based Metrics}}\\
         word embedding metrics & word lists & \citet{wambsganss-etal-2023-unraveling}; \citet{hansal-etal-2022-indigenous}\\
         sentence embedding metrics & word lists & \citet{sahoo-etal-2024-indibias}; \citet{malik-etal-2022-socially}\\
         \hline
         \multicolumn{2}{l}{\textbf{Probability-based Metrics}}\\
         masked token methods & counterfactual inputs \newline (masked) & \citet{vashishtha-etal-2023-evaluating}; \citet{guo-etal-2022-zhong}\\
         pseudo-log-likelihood methods & counterfactual inputs \newline (unmasked) & \citet{pikuliak-etal-2023-depth}; \citet{kaneko2022gender}\\
         \hline
         \multicolumn{2}{l}{\textbf{Generated Text-based Metrics}}\\
         distribution metrics & prompts & \citet{li-etal-2024-preference} (sentence completion); \citet{truong-etal-2024-crossing} (QA)\\
         classifier metrics & prompts & \citet{brun-nikoulina-2024-frenchtoxicityprompts} (sentence completion); \citet{mihaylov-shtedritski-2024-elegant} (QA)\\
         lexicon metrics & prompts & \citet{martinkova-etal-2023-measuring} (sentence completion); \citet{touileb-nozza-2022-measuring} (sentence completion)\\
         \hline
         \multicolumn{2}{l}{\textbf{Performance-based Metrics}}\\
         classification scores & classification instances & \citet{conti-wisniewski-2023-using}; \citet{huang-2022-easy}\\
         \hline\hline
    \end{tabular}
    \caption{Sample papers for each category of evaluation metrics and benchmark entry units used by multilingual bias studies, as grouped using typologies from \citet{gallegos-etal-2024-bias} and \citet{gupta-etal-2024-sociodemographic}. }
    \label{tab:bias_evaluation_metrics}
\end{table*}

In contrast, researchers working with heavily gendered languages faced the opposite conundrum: gender inflections often mutated minimally different inputs into almost completely different sentences. \citet{neveol-etal-2022-french} presents the case of the English pair \textit{\textbf{Women} talk a lot.} / \textit{\textbf{Men} talk a lot.} plausibly corresponding to \textit{Les \textbf{femmes} sont \textbf{bavardes}.} / Les \textbf{hommes} sont \textbf{bavards}. in French. These translations are problematic because half of the pair’s tokens are different from each other and will make the calculation of the bias metric unsound. The authors resorted to creative paraphrasing to circumvent the complication. For example, the above was translated into \textit{Les \textbf{femmes} parlent à tort et à travers.} / \textit{Les \textbf{hommes} parlent à tort et à travers.} which roughly translate to \textit{\textbf{Men} / \textbf{women} talk all over the place.}—preserving both the meaning and the minimal difference of the original English pair. 

A third issue was the duality of genders and meanings that a gendered word sometimes encoded in a language. In Icelandic, grammatically masculine nouns are generally used to refer to both male and female individuals despite feminine alternatives being sometimes present—for example, the masculine \textit{hjúkrunarfræðingur} is used to refer to Icelandic male and female nurses in spite of the feminine \textit{hjúkrunarkona} being available \citep{steinborn2022information,grigoreva-etal-2024-rubia}. These complexities make it hard to disentangle whether the biased model behaviors induced by these dually encoding words are linked to their inherent grammatical gender or to the multiple meanings they refer to in reality. As a result, one study eliminated the use of these words altogether and looked for reasonable substitutes instead \citep{chavez-mulsa-spanakis-2020-evaluating}. Others retained them but forewarned of their possible impact on evaluation results \citep{grigoreva-etal-2024-rubia,matthews-etal-2021-gender}. 

We end this subsection with our observation (Figure \ref{fig:findings}i) that most multilingual bias studies opt for gender proxies which represent only the male-female binary ($n=72, 67.92\%$) and fail to consider non-binary ($n=18, 16.98\%$) and transgender ($n=8,, 7.55\%$) identities. This propensity to ignore queerness is dangerous since it precludes work that can quantify and mitigate the harms PLMs can bring on non-heterosexual groups \citep{goldfarb-tarrant-etal-2023-prompt}. Given the contextually unique struggles of non-binary groups across different cultures \citep{Hinchy_2019,McMullin01012011,garcia1996phgay}, \textbf{we call on multilingual bias scholars to be more conscious not only in navigating linguistic gender features peculiar to their languages but also in actively pondering how they can incorporate the perspectives of queer communities in their cultures}. 

\section{Evaluation and Mitigation Methods}
\label{sec:eval_mitig}
Using the taxonomy of bias evaluation metrics proposed by \citet{gallegos-etal-2024-bias} and \citet{gupta-etal-2024-sociodemographic}, we found a relatively even mix of multilingual studies (Figure \ref{fig:findings}j, Table \ref{tab:bias_evaluation_metrics}) that measure bias in generated texts ($n=43, 40.57\%$), quantify bias based on comparing token probabilities ($n=35, 33.02\%$), and calculate bias using internal embedding vectors ($n = 32, 30.19\%$). This balance is mirrored in the studies’ benchmark formats of choice (Figure \ref{fig:findings}k): $30.08\%$ ($n=37$) use prompts often partnered with generated text-based metrics, $34.15\%$ ($n=42$) work with counterfactual sentence pairs frequently inputted into probability-based metric frameworks, and $29.27\%$ ($n=36$) involve word lists required for embedding-based metrics. We argue that this equilibrium in the kinds of bias evaluation approaches utilized in multilingual bias literature conceals a gap in the research area. \textbf{The fact that methods developed for word2vec and other static embeddings still constitute a significant proportion (about one-third) of multilingual bias research hints that the field has not yet fully caught up with Transformer-driven advancements in NLP}.\footnote{77.14\% of these embedding-based studies were conducted from 2020 onwards, indicating that they continue to be dominant despite the emergence and rapid development of multilingual Transformer-based models during this time.} Although the equally large number of studies operating on probability- and generated text-based metrics demonstrates progress and promise, more concerted efforts are still needed in ensuring the fairness and safety of the latest multilingual technologies deployed in non-English-speaking cultures. 

Equally notable in Figures \ref{fig:findings}j and \ref{fig:findings}k is the small number of studies employing performance-based metrics and benchmarks composed of classification instances ($n=8$). We expected this outcome because, as mentioned above, there are only a limited number of non-English labeled datasets that can be used in this respect \citep{joshi-etal-2020-state}. Thus, \textbf{there is also a need to devise benchmarks and studies that measure bias on downstream tasks in multilingual PLMs}. This direction of inquiry is critical, especially in light of some research suggesting the weak correlation between downstream model behavior and the probability- and embedding-based metrics currently dominating multilingual bias research \citep{cabello2023,delobelle-etal-2022-measuring}. 

This dearth in downstream multilingual bias research is matched by a scarcity of multilingual bias mitigation research as well (Figure \ref{fig:findings}l, Table \ref{tab:bias_mitig}). The overwhelming majority of the papers we looked into do not undertake bias mitigation experiments at all ($n=74, 69.81\%$). One possible reason for this is that many debiasing methods used for English models (e.g., data augmentation, instruction tuning, contrastive learning, adversarial learning) require readily available English datasets to balance biased pretraining data, to fine-tune existing models for fairness, or to modify their internal architectures (e.g., \citealp{zheng-etal-2023-click}; \citealp{zayed2023}; \citealp{narayanan-venkit-etal-2023-nationality}). \textbf{Such debiasing datasets are not easily accessible in non-English languages, making multilingual bias mitigation research scarce}. 

Among multilingual bias studies that do perform mitigation, pre-processing mitigation techniques are the most frequent ($n=28, 26.42\%$), with projection-based mitigation approaches ($n=11$) being the most widely used under this category. Projection-based mitigation identifies an embedding model subspace corresponding to a bias dimension (e.g., gender) and nullifies this subspace to minimize model bias \citep{gallegos-etal-2024-bias}. The prevalence of such a technique again signifies that much of the multilingual bias research still centers on an embedding-based language modeling framework. Consequently, we also deem as urgent the matter of updating and expanding endeavors to mitigate bias in multilingual PLMs.

\begin{table}[!t]
    \centering
    \small
    \begin{tabular}{ll}
        \hline
         \textbf{Mitigation Stage} & \textbf{Sample Papers}\\
         \hline
         pre-processing & \citet{ustun-etal-2024-aya}; \\ 
         & \citet{ahn2021mitigating}\\
         in-training & \citet{aakanksha-etal-2024-multilingual}; \\ 
         & \citet{ramesh-etal-2023-comparative}\\
         intra-processing & \makecell[l]{\citet{ermis-etal-2024-one}; \citet{lee-etal-2023-square}}\\
         post-processing & \citet{jain-etal-2022-leveraging}\\
        \hline
    \end{tabular}
    \caption{\small Sample studies that mitigate bias in multilingual models, as categorized by the stage in the language modeling pipeline at which they intervene.}
    \label{tab:bias_mitig}
\end{table}

\section{Conclusion and Future Directions}
\label{sec:conclusion}
In this paper, we sought to elucidate patterns and practices in the multilingual bias literature and to gauge their effectiveness in broadening the cultural inclusivity of PLM bias research. Our analysis uncovered opportunities for future research that can further accelerate the field’s growth. These opportunities include (but are not limited to):
\setlength{\itemsep}{0pt} 
\setlength{\topsep}{0pt}
\begin{itemize}
    \item evaluating social bias beyond cultures with high-resource and Indo-European languages to address \textit{linguistic bias} in multilingual PLM bias research ,
    \item employing culturally aware benchmark development methodologies that explicitly document cultural complexities,
    \item designing benchmarks that incorporate culturally specific bias dimensions and stereotypes collected from contextualized perspectives, 
    \item expanding research grounded on the heteronormative binary to include diverse expressions of queerness across cultures,
    \item pushing past embedding-based methods and reinforcing bias research on state-of-the-art multilingual models, especially those used in downstream tasks, and
    \item debiasing multilingual models.
\end{itemize}

We hope that researchers and practitioners working on multilingual bias can use our work to guide their own efforts to address bias in non-English contexts. We also hope that through our survey, they can leverage the myriad approaches which scholars around the world have taken to make PLMs safer and fairer for communities all over the globe.

\section*{Limitations}
Our work is subject to some limitations. First, a few of the papers we annotated were written in a non-English language. Specifically, \citet{guo-etal-2022-zhong} was written in Chinese while \citet{benamar-etal-2022-etude} was written in French. To allow us to include these in our survey, we used machine translators to translate the papers into English. This approach might have influenced the way we understood and annotated the papers. To minimize the impact of translation inaccuracies, we cross-referenced translations across different tools to confirm correctness. Furthermore, one of the authors has native proficiency in Mandarin while another has conversational proficiency; thus, they were capable of checking the translations for the Chinese article.

Second, while the categories and values we use in our taxonomy are based on previous PLM bias surveys, it is still possible that these do not encompass all extant or incoming research in the field.

Third, our search strategy did not include notable machine learning and artificial intelligence conferences (e.g., ICLR, ICML), nor did it consider non-English databases. However, it is interesting to note that most articles fulfilling our search criterion only come from ACL conferences. Despite including non-ACL venues (NeurIPS, FAccT, and AIES), only 2 papers from these conferences satisfied our criteria (1 from FAccT, 1 from AIES). This may suggest (although not conclusively) that multilingual bias studies rarely feature in non-ACL events. 

Finally, we focus on only the evaluation and mitigation aspects of the bias literature and do not examine research strands in the field that are only just emerging, such as explainability (e.g., \citealp{liu2024devil}; \citealp{conti-wisniewski-2023-using}) and interpretability (e.g., \citealp{gamboa-etal-2025-bias-attribution-fil}; \citealp{gamboa-2024-interpretability}).

We also acknowledge the potential psychosocial risks of compiling bias tests and benchmarks with possibly offensive entries into one location (i.e., the article annotation repository we share). However, we feel that the benefits of making these resources easily accessible (e.g., advancing multilingual bias research) outweighs such risks. 

%First, we only searched for candidate papers published or presented in ACL Anthology, IEEE Xplore, NeurIPS, FAccT, and AIES; thus, works matching our inclusion criteria but disseminated through other platforms were not included in our review. 

%Third, some aspects of the annotation taxonosmy required the authors to make subjective judgments about the papers examined. For example, assessment on whether a study’s methodological framework was culturally aware or not could have been affected by the authors’ personal viewpoints of what constitutes cultural sensitivity. However, as mentioned, disagreements were infrequent during annotation. 

%\section*{Ethical Considerations}
%We acknowledge that our personal experiences and assumptions inevitably shaped the way through which we evaluated the articles we surveyed and organized this paper’s narrative. Nevertheless, we strongly feel that this risk is minimal and outweighs the potential benefits and impact of this work. By doing a meta-analysis of how scholars have investigated the bias and ethical impact of language technologies on diverse cultural communities, we highlight trends, issues, and research directions that can expedite research on reducing the social harm of these NLP systems. 

\section*{Acknowledgments}
Lance Gamboa would like to thank the Philippine government's Department of Science and Technology for funding his doctorate studies.

\bibliography{custom}

\appendix

\section{Annotation Taxonomy}
\label{app:taxonomy}

\subsection{Linguistic Diversity}

\textbf{Language.} What non-English languages are considered?
\setlength{\itemsep}{0pt} 
\setlength{\topsep}{0pt}
\begin{itemize}
    \item \textit{languages listed in ISO 639} \citep{libraryofcongress2017iso}
\end{itemize}

\noindent \textbf{Language Family.} What language families do the non-English languages belong to?
\setlength{\itemsep}{0pt} 
\setlength{\topsep}{0pt}
\begin{itemize}
    \item Afro-Asiatic
    \item Altaic
    \item Austroasiatic
    \item Austronesian
    \item Dravidian
    \item Eskimo-Aleut
    \item Indo-European
    \item Japonic
    \item Kartvelian
    \item Koreanic
    \item Kra-Dai
    \item Niger-Congo
    \item Sino-Tibetan
    \item Uralic
\end{itemize}

\noindent\textbf{NLP Resources.} How much NLP resources do the languages have?
\setlength{\itemsep}{0pt} 
\setlength{\topsep}{0pt}
\begin{itemize}
    \item  Class 5 (most highly resourced, according to \citealp{joshi-etal-2020-state})
    \item Class 4
    \item Class 3
    \item Class 2
    \item Class 1
    \item Class 0 (lowest)
\end{itemize}

\subsection{Benchmark Development and Cultural Considerations}

\noindent\textbf{Benchmark Originality.} Is the evaluation benchmark original or adapted from an existing one?
\setlength{\itemsep}{0pt} 
\setlength{\topsep}{0pt}
\begin{itemize}
    \item original
    \item adapted
    \item mixed
\end{itemize}

\noindent\textbf{Benchmark Development Method.} How were the benchmark entries constructed?
\setlength{\itemsep}{0pt} 
\setlength{\topsep}{0pt}
\begin{itemize}
    \item written by authors
    \item contributed by annotators or experts
    \item crowdsourced
    \item machine-generated
    \item reused available benchmarks
\end{itemize}

\noindent\textbf{Number of Entries.} How many entries are in the benchmark?
\setlength{\itemsep}{0pt} 
\setlength{\topsep}{0pt}
\begin{itemize}
    \item \textit{integer value}
\end{itemize}

\noindent\textbf{Cultural Awareness.} How were cultural considerations in benchmark development documented?
\setlength{\itemsep}{0pt} 
\setlength{\topsep}{0pt}
\begin{itemize}
    \item explicit: The paper documents cultural nuances in detail.
    \item implicit: Cultural awareness is assumed through the participation of cultural insiders but not thoroughly detailed in paper.
    \item none: The paper shows little to no evidence of considering cultural nuances.
\end{itemize}

\noindent\textbf{Bias Dimension.} Which social dimensions are investigated demographic groups based on?
\setlength{\itemsep}{0pt} 
\setlength{\topsep}{0pt}
\begin{itemize}
    \item age
    \item caste
    \item criminal record
    \item culture
    \item disability
    \item disease
    \item education
    \item ethnicity
    \item family structure
    \item gender
    \item household registration
    \item immigration status
    \item intersectional
    \item marital status
    \item nationality
    \item occupation
    \item physical appearance
    \item politics
    \item pregnancy
    \item race
    \item region
    \item religion
    \item sexual orientation
    \item socioeconomic status
    \item unspecified
\end{itemize}

\noindent\textbf{Gender Proxy.} What terms are used to represent the gender groups being examined?
\setlength{\itemsep}{0pt} 
\setlength{\topsep}{0pt}
\begin{itemize}
    \item grammatical gender (e.g., \textit{the skilled engineering} translating to \textit{el inginiero experto} or \textit{la inginiera experta} in Spanish depending on gender)
    \item identity terms (e.g., \textit{male}/\textit{female})
    \item names (e.g., \textit{John}/\textit{Jane})
    \item pronouns (e.g., \textit{he}/\textit{she})
    \item roles (e.g., \textit{father}/\textit{mother})
    \item unspecified
\end{itemize}

\noindent\textbf{Gender Theorization.} How is gender conceptualized for papers examining gender bias?
\setlength{\itemsep}{0pt} 
\setlength{\topsep}{0pt}
\begin{itemize}
    \item binary only
    \item nonbinary-inclusive
    \item trans-inclusive
    \item undefined
\end{itemize}

\subsection{Bias Evaluation and Mitigation}

\noindent\textbf{Bias Evaluation Metric.} What metric is used to measure bias, as broadly classified according to the underlying data structure the metric operates on?
\setlength{\itemsep}{0pt} 
\setlength{\topsep}{0pt}
\begin{itemize}
    \item embedding-based metric
    \item generated text-based metric
    \item performance-based metric
    \item probability-based metric
    \item no bias evaluation
\end{itemize}

\noindent\textbf{Benchmark Entry Unit.} What format do benchmark entries follow?
\setlength{\itemsep}{0pt} 
\setlength{\topsep}{0pt}
\begin{itemize}
    \item classification instances
    \item counterfactual inputs – masked tokens
    \item counterfactual inputs – unmasked sentences
    \item prompts – question-answering
    \item prompts – sentence completions
    \item prompts – rewriting
    \item word pairs or lists
\end{itemize}

\noindent\textbf{Bias Mitigation Method (Level 1 Category).} What bias mitigation techniques are implemented, as broadly classified according to the LLM workflow stage at which they intervene?
\setlength{\itemsep}{0pt} 
\setlength{\topsep}{0pt}
\begin{itemize}
    \item pre-processing mitigation
    \item in-training mitigation
    \item intra-processing mitigation
    \item post-processing mitigation
    \item no bias mitigation
\end{itemize}

\noindent\textbf{Bias Mitigation Method (Level 2 Category).} What specific method is used to mitigate bias?
\setlength{\itemsep}{0pt} 
\setlength{\topsep}{0pt}
\begin{itemize}
    \item pre-processing mitigation: data augmentation, data filtering and reweighting, data generation, instruction tuning, projection-based mitigation, feature engineering
    \item in-training mitigation: architecture modification, loss function modification, selective parameter updating, filtering model parameters
    \item intra-processing mitigation: decoding strategy modification, weight redistribution, modular debiasing networks, bias reduction experts
    \item post-processing mitigation: rewriting, chain-of-thought
\end{itemize}

\section{Related Work}
\label{sec:related_work}

\subsection{PLM Bias Surveys}
\citet{blodgett-etal-2020-language} were among the first to organize the PLM bias literature into an organized meta-analysis. They borrowed the social sciences’ measurement modeling framework to unveil the tenuous ways by which bias studies in NLP conceptualize and operationalize bias. They follow up this work with another analysis exposing design flaws in widely utilized bias evaluation benchmarks, such as CrowS-Pairs, StereoSet, and WinoBias \citep{blodgett-etal-2021-stereotyping}. \citet{goldfarb-tarrant-etal-2023-prompt} continue this line of measurement modeling-based analyses by assessing the reliability and validity of ninety bias evaluation benchmarks—87\% of which are in English.

Other surveys of PLM bias include those carried out by \citet{czarnowska2021quantifying}, who categorized different fairness metrics into three groups, and \citet{10.1145/3597307}, who focused on the various social dimensions of bias explored by past studies. Most recently, \citet{gallegos-etal-2024-bias} and \citet{gupta-etal-2024-sociodemographic} separately published comprehensive typologies that were almost identical in their classification of bias evaluation metrics into embedding-based, probability-based, and generation-based measures and bias mitigation methods according to the stage in the training pipeline at which the mitigation intervention is administered. 

Our work’s objectives are most similar to the aims of \citet{Xu2025}, \citet{ramesh-etal-2023-fairness}, and \citet{talat2022reap}, who contemplate the difficulties of evaluating PLM bias multilingual and multicultural settings. Our analysis diverges from theirs in approach, in scope, and in the range of operationalization and method issues considered. While \citet{ramesh-etal-2023-fairness} include only seven multilingual datasets created mostly for text classification tasks, we inspect a bigger number of benchmarks for a broader variety of tasks. We also look beyond the research design factors they and \citet{Xu2025} concentrate on—language, bias dimension, evaluation metric, dataset task, and mitigation—and additionally highlight methodological elements linked to cultural awareness, adaptation methods, and gender theorization among others. On the other hand, the position paper by \citet{talat2022reap} reviews the field with a theoretically and conceptually dense perspective. We supplement this by juxtaposing their claims with the empirical evidence our systematic review collates.

\subsection{Cultural Awareness and Multilingual Benchmarks}
Most of the reviews discussed above call for the development of more multilingual and non-English bias benchmarks. NLP scholars from all over the globe have largely responded to this call (e.g., \citealp{lauscher-etal-2020-araweat}; \citealp{neveol-etal-2022-french}; \citealp{gamboa-lee-2025-filipino}); however, whether or not the benchmarks they developed are appropriate to the cultures of their chosen languages remains an unanswered question. Multilingual benchmarks used to assess PLMs often arise from machine translations of English language understanding benchmarks (e.g., multilingual MMLU used for GPT-4 in \citealp{openai2023gpt} and Llama 3 in \citealp{meta2024llama3}). Consequently, they not only suffer from quality issues but also fail to check and account for knowledge and nuances specific to the culture/s of the translated benchmark’s target language/s \citep{wibowo-etal-2024-copal,hershcovich-etal-2022-challenges}. These concerns are especially relevant in the field of PLM bias because values and stereotypes differ across cultures and countries \citep{talat2022reap}. For example, Korean and American cultures seem to differ in the way they stereotypically associate socioeconomic status with drug use: while an American bias test links drug use to impoverished individuals \citep{parrish2022bbq}, Korean researchers note that the reverse is true in their culture and that drug use is seen to be a pastime among the higher social classes of Korea \citep{jin-etal-2024-kobbq}. The intricacies of constructing culturally sensitive multilingual bias benchmarks are further affirmed by acknowledgments from benchmark creators themselves that their tests might be limited in scope and miss out some important stereotypes in their cultures (e.g., \citealp{sahoo-etal-2024-indibias}; \citealp{hsieh-etal-2024-twbias}). These complexities underscore the need to review the challenges faced and approaches taken by multilingual PLM bias studies in order to guide future research.

\section{Languages of Annotated Bias Evaluation Benchmarks}
\label{app:languages}

\begin{sidewaysfigure*}[!b]
\centering
  \includegraphics[width=\textwidth]{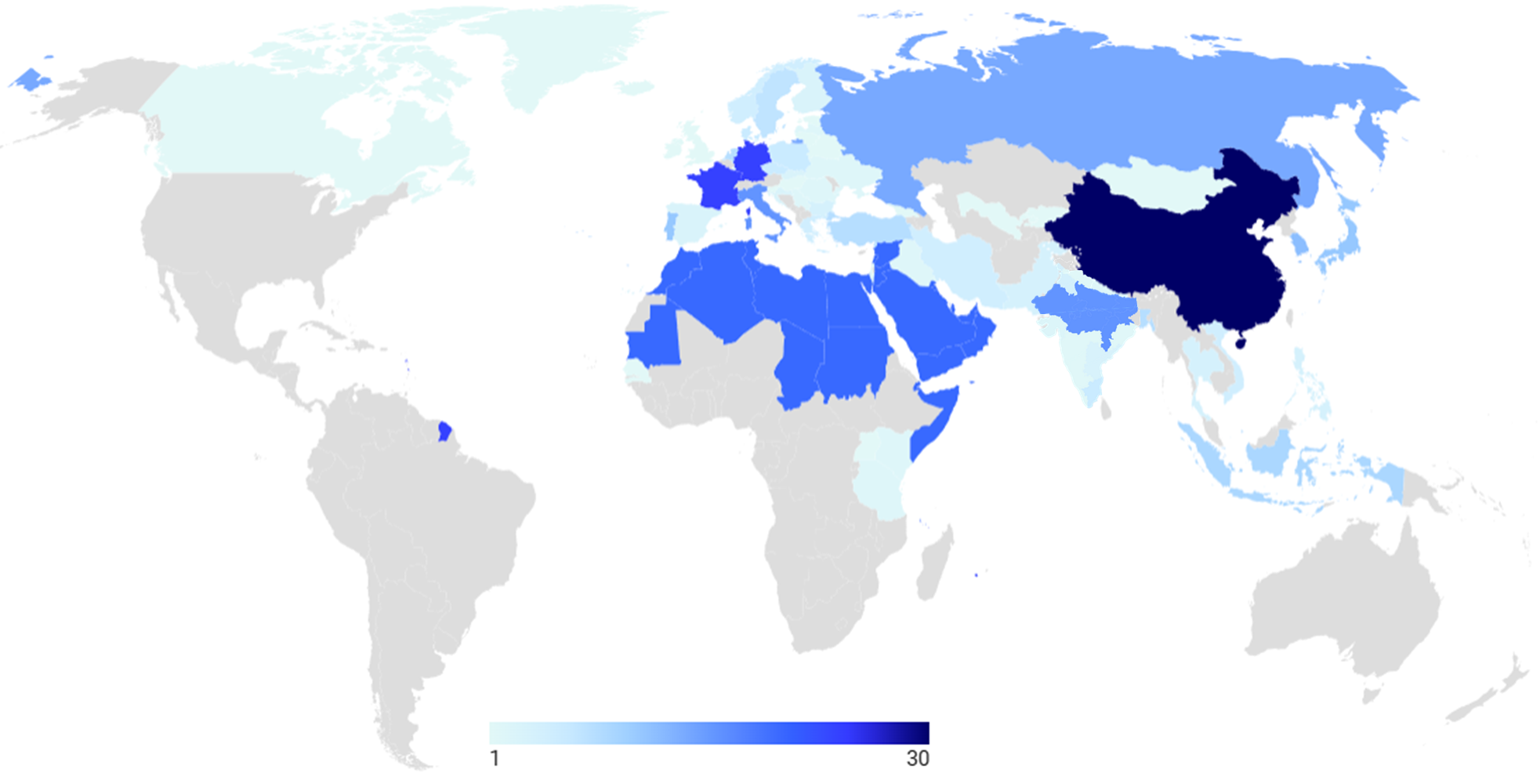}
  \caption{Regional heatmap of monolingual sub-benchmarks. The colors represent the number of benchmarks written using languages widely spoken in a particular country or territory.}
  \label{fig:heatmap}
\end{sidewaysfigure*}

See Table \ref{tab:languages} and Figure \ref{fig:heatmap}.

\begin{table*}[b]
    \centering
    \renewcommand{\arraystretch}{1.2} % Adjust row height for readability
    \setlength{\tabcolsep}{8pt} % Adjust column spacing
    \begin{tabular}{l c l c l c}
        \toprule
        \textbf{Language} & \textbf{$n$} & \textbf{Language} & \textbf{$n$} & \textbf{Language} & \textbf{$n$} \\
        \midrule
        Chinese & 30 & Persian & 5 & Assamese & 1 \\
        Spanish & 28 & Vietnamese & 5 & Belarusian & 1 \\
        French & 24 & Bulgarian & 4 & Estonian & 1 \\
        German & 24 & Filipino & 4 & Ganda & 1 \\
        Arabic & 20 & Punjabi & 4 & Georgian & 1 \\
        Italian & 16 & Thai & 4 & Hungarian & 1 \\
        Hindi & 16 & Urdu & 4 & Icelandic & 1 \\
        Russian & 14 & Catalan & 3 & Inuktitut & 1 \\
        Korean & 12 & Croatian & 3 & Irish & 1 \\
        Japanese & 11 & Finnish & 3 & Kannada & 1 \\
        Portuguese & 11 & Slovak & 3 & Konkani & 1 \\
        Indonesian & 9 & Telugu & 3 & Kyrgyz & 1 \\
        Bengali & 8 & Gujarati & 2 & Lithuanian & 1 \\
        Dutch & 8 & Hebrew & 2 & Luxembourgish & 1 \\
        Turkish & 8 & Kurdish & 2 & Mongolian & 1 \\
        Marathi & 7 & Latvian & 2 & Odia & 1 \\
        Swedish & 7 & Malayalam & 3 & Sanskrit & 1 \\
        Czech & 6 & Maltese & 2 & Slovenian & 1 \\
        Danish & 6 & Nepali & 2 & Uzbek & 1 \\
        Polish & 6 & Romanian & 2 & Welsh & 1 \\
        Tamil & 6 & Serbian & 2 & Wolof & 1 \\
        Greek & 5 & Swahili & 2 & & \\
        Norwegian & 5 & Ukrainian & 2 & & \\
        \bottomrule
    \end{tabular}
    \caption{Number of monolingual sub-benchmarks per language.}
    \label{tab:languages}
\end{table*}

\section{Sample Papers}
\label{app:sample_papers}
See Table \ref{tab:bias_dimensions}.

\begin{table*}[!htb]
    \centering
    \small
    \begin{tabular}{ll}
        \hline
         \textbf{Dimension} & \textbf{Sample Papers}\\
         \hline
         gender & \citet{bhutani-etal-2024-seegull}; \citet{demidova-etal-2024-john}\\
         religion & \citet{almazrouei-etal-2023-alghafa}; \citet{levy-etal-2023-comparing}\\
         race & \citet{nie-etal-2024-multilingual}; \citet{huang-etal-2024-flames}\\
         sexual orientation & \citet{bergstrand-gamback-2024-detecting}; \citet{mukherjee-etal-2023-global} \\
         age & \citet{10.5555/3716662.3716803}; \citet{neveol-etal-2022-french}\\
         nationality & \citet{zhu-etal-2024-quite}; \citet{das-etal-2023-toward}\\
         ethnicity & \citet{ramesh-etal-2023-comparative}; \citet{camara-etal-2022-mapping}\\
         disability & \citet{mina-etal-2024-exploring}; \citet{fort-etal-2024-stereotypical}\\
         physical appearance & \citet{zhao-etal-2023-chbias}; \citet{costa-jussa-etal-2023-multilingual}\\
         socioeconomic status & \citet{nie-etal-2024-multilingual}; \citet{grigoreva-etal-2024-rubia}\\
         region & \citet{billah-nagoudi-etal-2023-jasmine}; \citet{deng-etal-2022-cold}\\
         politics & \citet{al-ali-libovicky-2024-gender}; \citet{barkhordar-etal-2024-unexpected}\\
         intersectional bias & \citet{sahoo-etal-2024-indibias}; \citet{devinney-etal-2024-dont}\\
         caste & \citet{b-etal-2022-casteism}; \citet{bhatt-etal-2022-contextualizing}\\
         culture & \citet{naous-etal-2024-beer}; \citet{demidova-etal-2024-john}\\
         education & \citet{huang-xiong-2024-cbbq}; \citet{jin-etal-2024-kobbq} \\
         occupation & \citet{lee-etal-2024-exploring-inherent}; \citet{zhou-etal-2022-towards-identifying}\\
         immigration statu & \citet{ousidhoum-etal-2021-probing}; \citet{mukherjee-etal-2023-global}\\
         family structure & \citet{jin-etal-2024-kobbq}; \citet{lee-etal-2023-kosbi}\\
         marital status & \citet{lee-etal-2023-kosbi}\\
         criminal record & \citet{lee-etal-2023-kosbi}\\
         pregnancy  & \citet{lee-etal-2023-kosbi}\\
         household registration  & \citet{huang-xiong-2024-cbbq}\\
        disease & \citet{huang-xiong-2024-cbbq}\\
        \hline
    \end{tabular}
    \caption{Social dimensions analyzed by multilingual bias studies.}
    \label{tab:bias_dimensions}
\end{table*}

\end{document}